%% file: main.tex
\documentclass{article}

\usepackage{microtype}
\usepackage{graphicx}
\usepackage{booktabs} 
\usepackage{subcaption}
\usepackage{tikzducks}
\usepackage{adjustbox}
\usepackage{glossaries}

\usepackage{hyperref}

\usepackage[accepted]{icml2024}

\usepackage{amsmath}
\usepackage{amssymb}
\usepackage{mathtools}
\usepackage{amsthm}

\usepackage{comment}
\usepackage{todonotes}

\glsdisablehyper
\loadglsentries{glossary}

\usepackage[capitalize,noabbrev]{cleveref}

\theoremstyle{plain}

\theoremstyle{definition}

\theoremstyle{remark}

\DeclareMathOperator{\mean}{mean} 

\icmltitlerunning{Self-supervised learning for crystal property prediction via denoising}

\begin{document}

\twocolumn[
\icmltitle{Self-supervised learning for crystal property prediction via denoising}



\icmlsetsymbol{equal}{*}

\begin{icmlauthorlist}
\icmlauthor{Alexander New}{xxx}
\icmlauthor{Nam Q. Le}{xxx}
\icmlauthor{Michael J. Pekala}{xxx}
\icmlauthor{Christopher D. Stiles}{xxx}
\end{icmlauthorlist}

\icmlaffiliation{xxx}{Research and Exploratory Development Department, Johns Hopkins University Applied Physics Laboratory, 11100 Johns Hopkins Rd, Laurel, MD 20723, USA}

\icmlcorrespondingauthor{Alexander New}{alex.new@jhuapl.edu}

\icmlkeywords{Machine Learning, ICML}

\vskip 0.3in
]



\printAffiliationsAndNotice{}  

\begin{abstract}
Accurate prediction of the properties of crystalline materials is crucial for targeted discovery, and this prediction is increasingly done with data-driven models. However, for many properties of interest, the number of materials for which a specific property has been determined is much smaller than the number of known materials. To overcome this disparity, we propose a novel \gls{SSL} strategy for material property prediction. Our approach, \gls{CDSSL}, pretrains predictive models (e.g., graph networks) with a pretext task based on recovering valid material structures when given perturbed versions of these structures. We demonstrate that \gls{CDSSL} models out-perform models trained without \gls{SSL}, across material types, properties, and dataset sizes.
\end{abstract}

\glsresetall

\section{Introduction}\label{sec:introduction}

Recent years have seen the development of efficient and accurate \gls{ML} methods for predicting properties of crystalline materials using descriptors based on composition~\cite{Ward2016magpie,Goodall2020roost,Wang2021crabnet,Pogue2023closedloop} and structure~\cite{Xie2018CGCNN,chen2019megnet,Choudhary2021alignn,New2022curvatureinformed,Ruff2024cogn}. These methods have demonstrated success across different material classes and properties~\cite{Dunn2020matbench}. They typically rely on \glspl{GN}~\cite{Battaglia2018graphnetworks}, in which nodes are atom, and edges capture inter-atom distances.

However, for many properties of interest, the number of material structures for which a property value is known is much less than the total number of stable materials that have a known structure. For example, the \gls{NOMAD} computational database~\cite{Scheidgen2023nomad} contains more than three million materials, and the \gls{OQMD}~\cite{Kirklin2015OQMD} contains more than one million. However, the shear modulus dataset in MatBench~\cite{Dunn2020matbench} contains only ten thousand materials. This disparity in relative sizes will only increase as generative models are increasingly used to predict novel material structures~\cite{Xie2022cdvae,Zhao2023pgcgm,new2023evaluating,Zeni2024mattergen}.

In order to make use of these large general-purpose databases and to avoid needing to invest time and effort in annotation for properties with little data, a natural solution is \gls{SSL}~\cite{Balestriero2023sslcookbook}. Unlike traditional \gls{SL}, in \gls{SSL}, models are trained on pretext tasks without need for labels, and then they are fine-tuned on the prediction task of interest. \gls{SSL} has been used widely in conjunction with \glspl{GN}~\cite{Xie2023graphsslreview}, especially in the context of molecular property prediction~\cite{Hu2020PretrainingStrategies,Godwin2022simple}. Some work has also used \gls{SSL} for crystalline material property prediction~\cite{Magar2022crystaltwin,Huang2024compositionssl}.

Zaidi \emph{et al.} recently developed a novel \gls{SSL} method for molecular property prediction based on structure-denoising~\cite{Zaidi2023denoising}. In particular, they showed that perturbing the atom positions in a molecule with noise and then training a model to predict that noise corresponded to learning an approximate force field for that molecule. This enabled accurate predictions of varied molecular properties.

\begin{figure*}
    \centering
    \includegraphics[width=\linewidth]{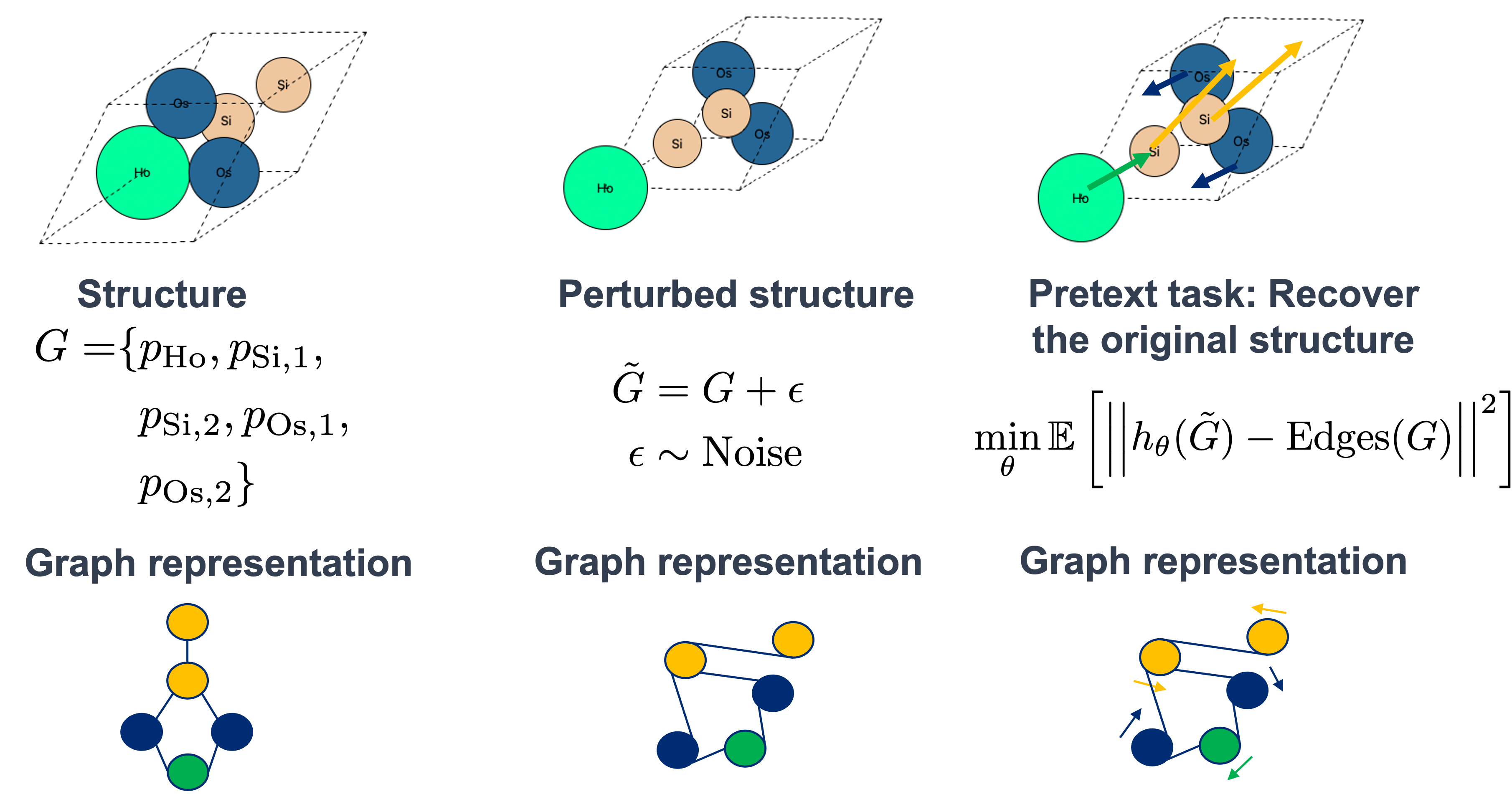}
    \caption{We summarize \gls{CDSSL}. The node positions of a structure $G$ are perturbed with Gaussian noise to create a structure $\tilde{G}$. The \gls{ML} model $h_\theta$ takes the perturbed structure  $\tilde{G}$ as input and seeks to output the edge embeddings of the original structure $G$.}
    \label{fig:ssl_schematic}
\end{figure*}

In this work, we develop a similar denoising \gls{SSL} approach  for crystalline structures. Our method, \gls{CDSSL}, works by perturbing the position of atoms in a material structure multigraph (\cref{sec:data} and~\cref{fig:ssl_schematic}) and then trains a model to predict the original structure's inter-atom distances (\cref{sec:cdssl}). In~\cref{sec:application}, we demonstrate how to combine \gls{CDSSL} with crystal property prediction models for specific prediction tasks. In~\cref{sec:results}, we evaluate \gls{CDSSL}, including assessments that vary the amount of training data, the material class of interest, and the target property. We show that \gls{CDSSL} consistently yields more accurate property-prediction models than those only using \gls{SL}. In~\cref{sec:assessment}, we demonstrate that the \gls{CDSSL} representation space captures some variation in properties even without finetuning.

\section{Methods}\label{sec:methods}

\subsection{Multigraph representations of materials}\label{sec:data}

Let $\mathcal{M}$ be a space of material crystal structures. We follow the \gls{CGCNN}~\cite{Xie2018CGCNN} approach and represent materials $g$ as directed multigraphs $G = (V, E)$ consisting of sets of nodes $V = \{v\}$ and edges $E = \{(v,v',k)\}$. Each node $v$ has a node embedding $x_v \in \mathbb{R}^{d_V}$ and a Cartesian position vector $p_v\in\mathbb{R}^3$, and each edge $(v,v',k)$ has an edge embedding $e_{v,v',k} \in \mathbb{R}$. Because $G$ describes a periodic tiling of a unit cell, a pair of nodes $v$ and $v'$ may be connected by multiple edges, indexed by $k = 1,\hdots,K_{v,v'}$. 

In the \gls{CGCNN} multigraph construction, the edge embedding $e_{v,v',k}$ reflects the distance between $v$ and $v'$: it is a function of both the node positions $p_v$ and $p_{v'}$ and the edge index $k$. Given a material structure, edges $(v,v',k)$ are constructed using nearest-neighbor calculations based on a given structure's lattice. See~\cref{tab:data_parameters} in~\cref{sec:hyperparameters} for further details on the multigraph representation.

\subsection{Crystal denoising self-supervised learning}\label{sec:cdssl}

\Cref{fig:ssl_schematic} outlines \gls{CDSSL}. Given a graph $G$, we generate a perturbed version $\tilde{G}$ of it by perturbing each node's position with Gaussian noise $\tilde{p}_v\sim\mathcal{N}(p_v, \sigma^2 I)$, for some variance $\sigma^2$. Note that, compared to $G$, $\tilde{G}$ has the same nodes, node embeddings, and edges, but different node positions and edge embeddings.

Let $h_\theta:G\mapsto\{y_{v,v',k}\}$ be a \gls{NN}, parameterized by a vector $\theta$, that maps a graph $G$ to a set of scalars $\{\hat{e}_{v,v',k}\}$, one for each of $G$'s edges. Then we define the \gls{CDSSL} pretext task as the following minimization problem:
\begin{eqnarray}
    \hat{\theta} &=& \arg\min_\theta \mathbb{E}_{p(\tilde{G}|G)p(G)}\left[\left|\left|h_\theta(\tilde{G}) - \bar{E}\right|\right|^2\right].
    \label{eq:ssl_task}
\end{eqnarray}
Here, $p(G)$ samples graphs $G$ from the training set, $p(\tilde{G}|G)$ generates perturbed graphs, the loss is calculated as $||h_\theta(\tilde{G}) - \bar{E}||^2 = \sum_{v,v',k}|\hat{e}_{v,v',k} - \bar{e}_{v,v',k}|^2$, and $\bar{E}$ is the set of normalized edge embeddings for $G$:
\begin{eqnarray}
    \bar{E} &=& \{\bar{e}_{v,v',k}\} = \left\{\frac{e_{v,v',k} - \mathrm{mean}\left\{e_{v,v',k}\right\}}{\mathrm{std}\left\{ e_{v,v',k}\right\}}\right\}.
    \label{eq:normalized_edges}
\end{eqnarray}
When every edge of a graph $G$ has the same embedding, we set each entry of $\bar{E}$ to $0$.

We present an interpretation of eq.~\ref{eq:ssl_task}. If the training set consists of structures at equilibrium, then the perturbation moves them away from locally minimizing the potential energy distribution. Thus, a model $h_\theta$ that solves eq.~\ref{eq:ssl_task} has learned to identify small shifts $\hat{e}_{v,v',k}$ that move a non-equlibrium structure into equilibrium. As has been argued in previous work for non-periodic molecules~\cite{Zaidi2023denoising}, learning this task of predicting equilibrium structures for arbitrary materials is equivalent to learning to minimize general interatomic potential functions. This justifies why we expect such an $h_\theta$ to have learned a general-purpose representation of materials space.

The \gls{CDSSL} noise hyperparameter scale $\sigma$ requires tuning. If it is too small, then the perturbed edge distances $\hat{e}_{v,v',k}$ are too similar to the original edge distances $e_{v,v',k}$, and \gls{CDSSL} pretraining objective (eq.~\ref{eq:ssl_task}) is minimized when the network $h_\theta$ memorizes the training set structures. If $\sigma$ is too large, then the perturbed $\tilde{G}$ is too different from the original structure $G$ for the objective to be minimizable. Here, we show results only for a single value of $\sigma$ and leave further exploration to future work.

The construction of the \gls{CDSSL} pretext task is independent of the precise form of $h_\theta$. All that is required of $h_\theta$ is that it can ingest node and/or edge embeddings and output per-edge quantities. This means that \gls{CDSSL} can be used in conjunction with general structure-based property prediction architectures, such as \glspl{CGCNN}~\cite{Xie2018CGCNN}, \glspl{MEGNET}~\cite{chen2019megnet}, \glspl{M3GNET}~\cite{Chen2022m3gnet}, \glspl{ALIGNN}~\cite{Choudhary2021alignn}, \glspl{CHGNET}~\cite{Deng2023Chgnet}, or others.

\subsection{Crystal denoising with MEGNets}\label{sec:application}

\begin{figure*}
    \centering
    \includegraphics[width=\linewidth]{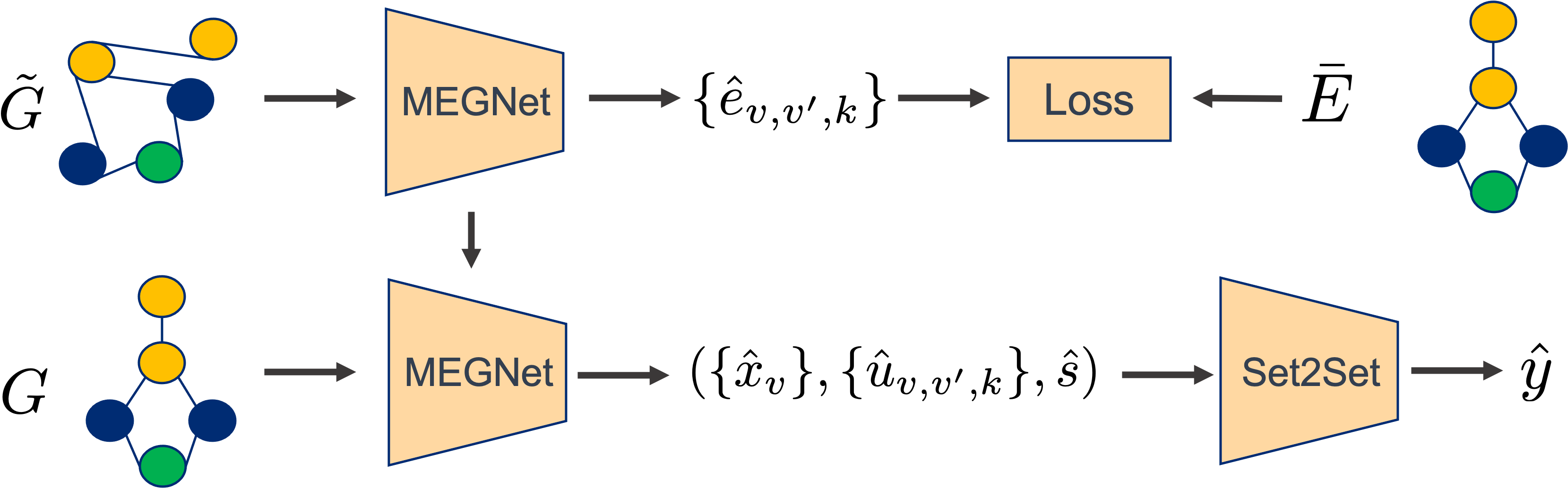}
    \caption{We summarize the application of our \gls{CDSSL} framework to a property prediction task. In the top row, a \gls{MEGNET} is trained to denoise crystal structures (\cref{fig:ssl_schematic} and eq.~\ref{eq:ssl_task}) with predicted edge embeddings $\hat{e}_{v,v',k}$. Once the \gls{MEGNET} module has been trained, we can finetune it on property prediction. This entails passing the node-level outputs $\hat{x}_v$, edge-level outputs $\hat{u}_{v,v',k}$, and graph-level output $\hat{s}$ through Set2Set~\cite{Vinyals2016set2set} modules to output the predicted property $\hat{y}$.}
    \label{fig:prediction_transfer}
\end{figure*}

In this work, we focus on using \glspl{MEGNET} as the base for $h_\theta$. We summarize our approach in~\cref{fig:prediction_transfer}. The \gls{MEGNET} graph convolution is defined by:
\begin{eqnarray}
    (\{x_v\}_v, \{e_{v,v',k}\})&\mapsto&(\{\hat{x}_v\},\{\hat{u}_{v,v',k}\},\hat{s}),
    \label{eq:megnet}
\end{eqnarray}
where $\hat{x}_v$ are node-level output vectors, $\hat{u}_{v,v',k}$ are edge-level output vectors, and $\hat{s}$ is a graph-level output vector. During pretraining, we map edge-level output vectors to predicted edge embeddings $\hat{e}_{v,v',k}$ with a linear layer:
\begin{eqnarray}
    \hat{e}_{v,v',k} &=& \mathrm{Linear}(u_{v,v',k}).
    \label{eq:distance_prediction}
\end{eqnarray}
When finetuning a pretrained model for a property prediction task, we follow typical practice for \glspl{MEGNET}  and use Set2Set~\cite{Vinyals2016set2set} modules to aggregate $\{\hat{x}_v\}$ and $\{\hat{u}_{v,v',k}\}$ into single vectors, and then we predict properties $\hat{y}$ with \glspl{MLP}:
\begin{eqnarray}
    \hat{y} &=& \mathrm{MLP}(\mathrm{Set2Set}(\{\hat{x}_v\}), \mathrm{Set2Set}(\{\hat{u}_{v,v',k}\}), \hat{s}).\,\,\,\,\,\,\,
\end{eqnarray}
We jointly train the \gls{MEGNET} module, the Set2Set modules, and \gls{MLP} for a target property, using the \gls{MSE} of standardized property values as the loss.

\section{Results}\label{sec:results_general}

\subsection{Evaluation details}\label{sec:evaluation}

We rely on \gls{MGL}~\cite{Ko2021matgl} for general data ingestion and loading procedures and for the \gls{MEGNET}~\cite{chen2019megnet} implementation. In particular, this leverages \texttt{pymatgen}~\cite{Ong2013pymatgen} to ingest \glspl{CIF}. The \texttt{pymatgen} structures are then converted into \gls{DGL}~\cite{Wang2020dgl} multigraphs. 

We evaluate \gls{CDSSL} on a variety of scalar regression material property-prediction tasks provided by MatMiner~\cite{Ward2018matminer}\footnote{\url{https://hackingmaterials.lbl.gov/matminer/dataset_summary.html}}. The dataset, dataset size, and target property are given in~\cref{tab:datasets}. 
We use the matbench\_mp\_e\_form dataset (which contains $132{,}752$ structures) as the training set for pretraining with \gls{CDSSL}. Further details on the datasets are in~\cref{tab:dataset_details} in~\cref{sec:hyperparameters}.

All hyperparameters for pretraining and training are in~\cref{sec:hyperparameters}. We pretrain and train with Adam~\cite{Kingma2014adam}; \glspl{MEGNET} use $\mathrm{SoftPlus2}(x) = \log(\exp(x)+1) - \log(2)$ activation. We set the \gls{CDSSL} noise scale to $\sigma=0.5$, which we chose after experimentation using matbench\_log\_grvh as the pretraining and evaluation dataset. 

\gls{SSL} is especially relevant in the setting where the amount of available labeled data is very small. Thus, we evaluate \gls{CDSSL} in both low-data and high-data settings. Specifically, we vary the training dataset size to be between $10\%$ and $70\%$ of the total dataset size (in increments of $10\%$). 

\begin{table}[h]
    \centering
    \begin{adjustbox}{width=0.48\textwidth}
    \begin{tabular}{c|c|c}
    Dataset                 &   Size        &   Property \\\hline
    boltztrap\_mp           &   $8{,}924$   &   s\_n\\
    dielectric\_constant    &   $1{,}056$   &   log10(poly\_total)\\
    jarvis\_dft\_2d         &   $636$       &   exfoliation\_en\\
    matbench\_log\_gvrh     &   $10{,}987$  &   log10(G\_VRH)\\
    matbench\_log\_kvrh     &   $10{,}987$  &   log10(K\_VRH)\\
    matbench\_perovskites   &   $18{,}928$  &   e\_form\\
    matbench\_phonons       &   $1{,}265$   &   log10(last phdos peak)\\\hline
    matbench\_mp\_e\_form   &   $132{,}752$ &   Not used\\
    \end{tabular}
    \end{adjustbox}
    \caption{The dataset name, size, and target property used in our evaluation of \gls{CDSSL}. The matbench\_mp\_e\_form dataset is used only for pretraining and not for property prediction. Datasets range in size from less than a thousand structures to tens of thousands of structures, and target properties include mechanical, electronic, and thermodynamic quantities.}
    \label{tab:datasets}
\end{table}

\subsection{Evaluation results}\label{sec:results}

\begin{figure}[h]
    \centering
    \includegraphics[width=\linewidth]{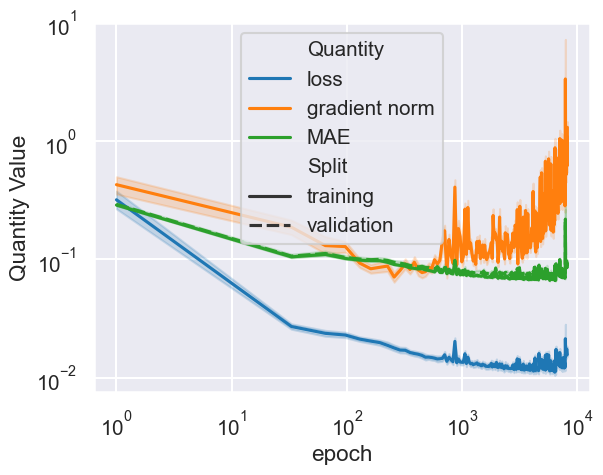}
    \caption{We show metrics from pretraining a \gls{MEGNET} with the \gls{CDSSL} pretraining objective. Training yields slow but consistent decreases in both the training loss (eq.~\ref{eq:ssl_task}) and the \gls{MAE} of the $h_\theta(\tilde{G}) - \bar{E}$ quantity (for both the training and validation set). \glspl{MAE} for the training and validation set overlap, indicating that overfitting is not happening. The \gls{CDSSL} pretraining task retains instabilities during training, as evidenced by the jump in metrics and gradient norm of the loss at the end of training.}
    \label{fig:ssl_traj}
\end{figure}

We use the matbench\_mp\_e\_form dataset for pretraining a \gls{MEGNET} with \gls{CDSSL}, with 80\% for training and 20\% for validation.  In~\cref{fig:ssl_traj}, we show that the \gls{CDSSL} pretraining task can be solved over the course of training and demonstrates no evidence of overfitting. However, it has some instability in later epochs. Thus, when finetuning \gls{CDSSL} models on \gls{SL} tasks, we identify the checkpoint that attained minimal pretraining loss and use that as the initial model.

\begin{figure*}[h]
    \centering
    \includegraphics[width=\linewidth]{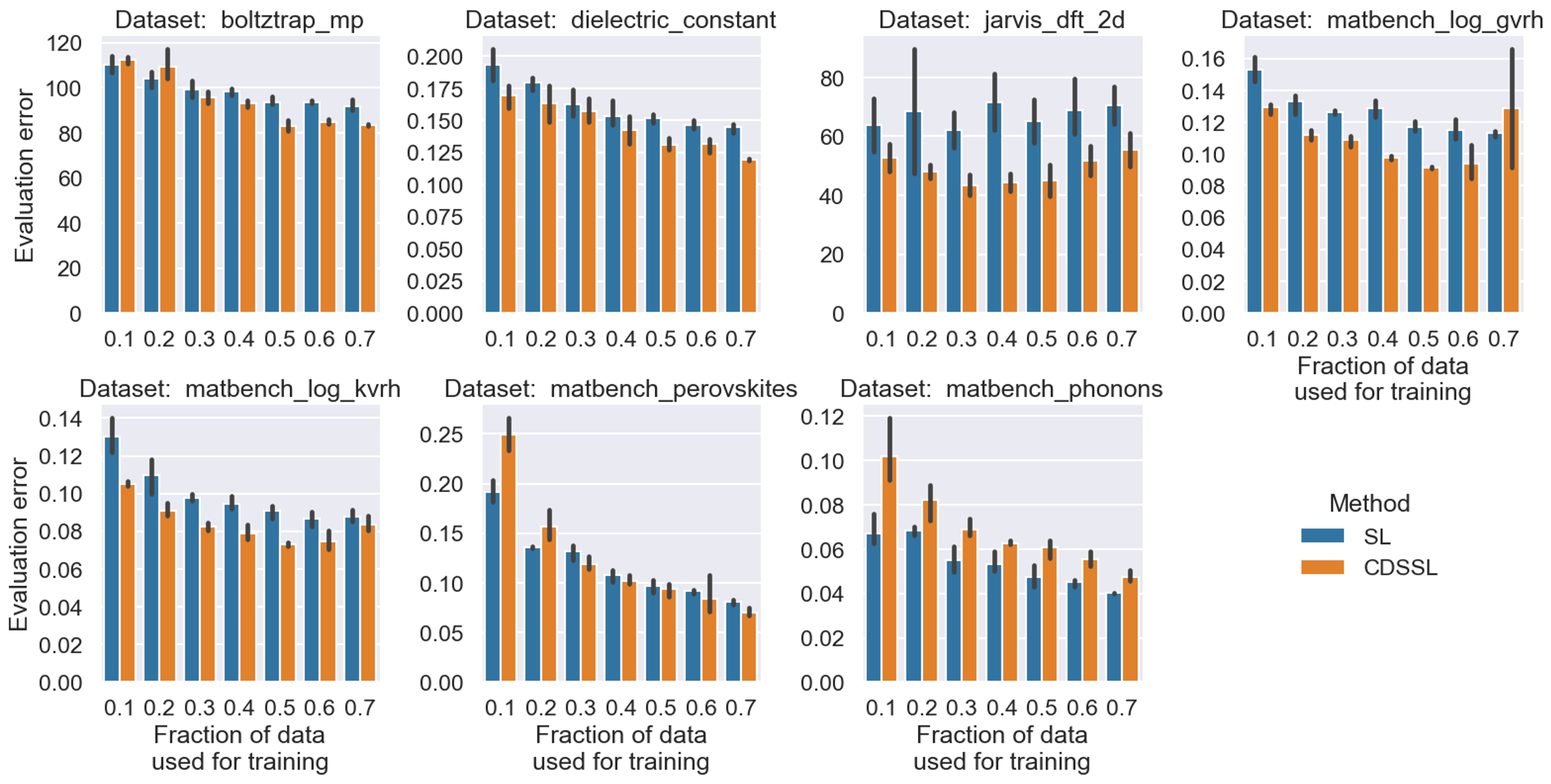}
    \caption{We demonstrate the effects of using \gls{CDSSL} vs. \gls{SL} across a variety of datasets and dataset sizes. Each bar reports error on the evaluation set, averaged over $3$ data splits and network initializations, and error bars show standard errors in estimating that mean accuracy. The model finetuned after \gls{CDSSL} has a lower error than the \gls{SL} model in $37$ out of $49$ (dataset, dataset size) configurations.}
    \label{fig:accuracy}
\end{figure*}

\Cref{fig:accuracy} shows the results of our study. In particular, \glspl{MEGNET} finetuned after pretraining with \gls{CDSSL} achieve lower evaluation error than \glspl{MEGNET} trained only with \gls{SL} in $37$ out of the $49$ (dataset, dataset size) configurations. This is an improvement in error across a wide variety of material classes, dataset sizes, and material property types. This suggests that a model pretrained with \gls{CDSSL} could be the basis for general-purpose material property prediction.

\subsection{Assessing the CDSSL representation space}\label{sec:assessment}

Our hypothesis for why \gls{CDSSL} works is that the pretext task (eq.~\ref{eq:ssl_task}) enables $h_\theta$ to learn a general representation of materials space. To test this hypothesis, we assess the quality of the \gls{CDSSL}'s learned representation for prediction tasks without additional fine-tuning. In particular, we choose $4{,}096$ points from the validation split of matbench\_mp\_e\_form and calculate their \gls{CDSSL} embeddings:
\begin{eqnarray}
    \hat{z} &=& \mathrm{Concat}
    \left(\mean_v\{\hat{x}_v\}, \mean_{v,v',k}\{\hat{u}_{v,v',k}\}, \hat{s}\right)\,\,\,\,\,\,\,\,\,\,\,
    \label{eq:representation}
\end{eqnarray}
where $\hat{x}$, $\hat{u}_{v,v',k}$, and $\hat{s}$ are the outputs of the \gls{MEGNET} graph convolution module, as in~\cref{sec:application}. 

We can estimate how informative the \gls{CDSSL} representation space is for material properties by using a linear probing strategy~\cite{Balestriero2023sslcookbook}. We use $80\%$ of these points to train ridge regressors to predict the log-transformed structure densities and volume, as calculated by \texttt{pymatgen}~\cite{Ong2013pymatgen}. We find that the density-prediction regressor attains an $R^2$ of $70.3\%$, and the volume-prediction regressor attains an $R^2$ of $75.6\%$. 

\begin{figure}
    \centering
    \includegraphics[width=\linewidth]{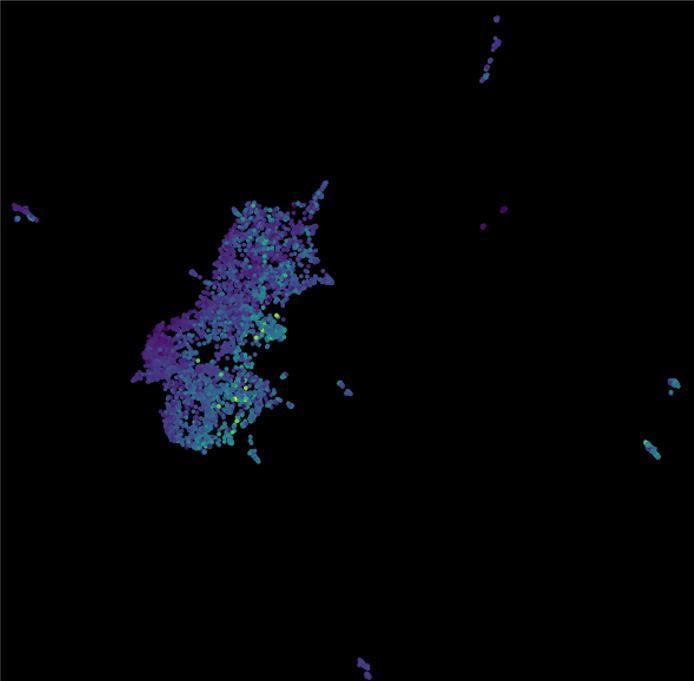}
    \caption{We use \gls{UMAP}~\cite{Mcinnes2020umap} to learn a reduced representation of the matbench\_mp\_e\_form dataset used for pretraining with \gls{CDSSL} (eq.~\ref{eq:representation}). We shade points by their corresponding structure's density. Within the reduced representation, structures with similar densities are near each other. This suggests that the representation space learned via \gls{CDSSL} has captured general notions of material properties. Error metrics are reported in the unit of each dataset's property.}
    \label{fig:umap}
\end{figure}

These results show that, even without the additional expressivity granted by \gls{MEGNET}'s $\mathrm{Set2Set}$ modules, the \gls{CDSSL} pretraining task enables models to learn representations of materials space. In~\cref{fig:umap}, we present additional evidence for this claim, where we use \gls{UMAP}~\cite{Mcinnes2020umap} to visually show that the \gls{CDSSL} space captures variation in material density.

\section{Discussion}\label{sec:discussion}

In this work, we have introduced a novel method, \gls{CDSSL}, for pretraining material property-prediction models. Our method works by taking a crystal structure, perturbing the positions of its constituent atoms with noise, and then tasking the predictive model to recover the structure's original edge embeddings. This enables the predictive model to learn a general, property-agnostic representation of material structure space. \gls{CDSSL} is generally applicable to structure-based property prediction models, but here we focused on the \gls{MEGNET}~\cite{chen2019megnet} architecture.

We showed that using \gls{CDSSL} for pretraining \glspl{MEGNET} yielded an increase in accuracy across a variety of datasets and material properties, compared to a \gls{MEGNET} trained only with \gls{SL}. However, we believe further work can enhance the effectiveness of \gls{CDSSL}. In particular, modification of the \gls{CDSSL} training loss might make its minimization process more stable. Such modification might come from the development of theory that can rigorously link the denoising task to statistical mechanics.

\section*{Acknowledgements}

This work was supported by internal research and development funding from the Research and Exploratory Development Mission Area of the Johns Hopkins University Applied Physics Laboratory.


\clearpage
\bibliography{references}
\bibliographystyle{icml2024}

\newpage
\appendix

\section{Supplemental data}\label{sec:hyperparameters}

\begin{table}[h]
    \centering
    \begin{tabular}{c|c}
    Hyperparameter                      &   Value\\\hline
    Perturbation scale $\sigma$         &   $0.5$\\
    hidden\_ layer\_ sizes\_input       &   $(128, 64)$\\
    hidden\_layer\_sizes\_conv          &   $(128, 128, 64)$\\
    hidden\_layer\_sizes\_output        &   $(64, 32)$\\
    dim\_node\_embedding                &   $16$\\
    dim\_edge\_embedding                &   $100$\\
    dim\_state\_embedding               &   $2$\\
    n\_blocks                           &   $3$ \\
    nlayers\_set2set                    &   $1$ \\
    niters\_set2set                     &   $2$ \\
    Optimizer                           &   Adam\\
    Activation                          &   SoftPlus2\\
    Minibatch size                      &   $256$\\
    Number of epochs                    &   $4096$
    \end{tabular}
    \caption{Hyperparameters used when pretraining models. See the \gls{MGL}~\cite{Ko2021matgl} documentation for details on what \gls{MEGNET}-specific hyperparameters mean.}
    \label{tab:ssl_hyperparameters}
\end{table}

\begin{table}[h]
    \centering
    \begin{tabular}{c|c}
    Hyperparameter                      &   Value\\\hline
    hidden\_ layer\_ sizes\_input       &   $(64, 32)$\\
    hidden\_layer\_sizes\_conv          &   $(64, 64, 32)$\\
    hidden\_layer\_sizes\_output        &   $(32, 16)$\\
    dim\_node\_embedding                &   $16$\\
    dim\_edge\_embedding                &   $100$\\
    dim\_state\_embedding               &   $2$\\
    n\_blocks                           &   $3$ \\
    nlayers\_set2set                    &   $1$ \\
    niters\_set2set                     &   $2$ \\
    \end{tabular}
    \caption{Hyperparameters used for the \gls{MEGNET} model trained from scratch. See the \gls{MGL}~\cite{Ko2021matgl} documentation for details on what \gls{MEGNET}-specific hyperparameters mean.}
    \label{tab:from_scratch_hyperparameters}
\end{table}

\begin{table}[ht!]
    \centering
    \begin{tabular}{c|c}
    Hyperparameter      &   Value \\\hline
    Optimizer           &   Adam\\
    Activation          &   SoftPlus2\\
    Number of epochs    &   $96$\\
    Learning rate       &   $1e-3$\\
    Minibatch size      &   $128$\\
    \end{tabular}
    \caption{Hyperparameters used both by the \gls{MEGNET} model trained from scratch and the model finetuned after pretraining.}
    \label{tab:shared_hyperparameters}
\end{table}

\begin{table}[ht!]
    \centering
    \begin{tabular}{c|c}
    Parameter                                   &   Setting\\\hline
    Cutoff distance for edge construction       &   5 \\
    \# of centers in Gaussian radial expansion  &   100\\
    Width of Gaussian functions                 &   0.5\\
    \end{tabular}
    \caption{Parameters used when constructing multigraph representations of material structures.}
    \label{tab:data_parameters}
\end{table}

\begin{table*}[t!]
    \centering
    \begin{tabular}{c|c|c}
    Dataset                 &       Property        &   Property description\\\hline
    boltztrap\_mp           &       s\_n            &   $n$-type Seebeck coefficient\\ 
    dielectric\_constant    &       poly\_total     &   average of eigenvalues of total contributions to the dielectric tensor\\
    jarvis\_dft\_2d         &       exfoliation\_en         &   exfoliation energy\\
    matbench\_log\_grvh     &       g\_vrh          &   Voigt-Reuss-Hill average shear modulus\\
    matbench\_log\_kvrh     &       k\_vrh          &   Voigt-Reuss-Hill average bulk modulus\\
    matbench\_perovskites   &       e\_form         &   Heat of formation\\
    matbench\_phonons       &   last\_phdos\_peak   &   Frequency of the highest frequency optical phonon mode peak
    \end{tabular}
    \caption{The target property for each dataset used in our studies. More details are available at MatMiner~\cite{Ward2018matminer} (\url{https://hackingmaterials.lbl.gov/matminer/dataset_summary.html}).}
    \label{tab:dataset_details}
\end{table*}






\end{document}

%% file: glossary.tex
\newacronym{APL}{APL}{%
  the Johns Hopkins University Applied Physics Laboratory%
}
\newacronym{ALIGNN}{ALIGNN}{atomistic line graph neural network}


\newacronym{CIF}{CIF}{crystallographic information file}
\newacronym{CV}{CV}{computer vision}
\newacronym{CGCNN}{CGCNN}{crystal graph convolutional neural network}
\newacronym{CDSSL}{CDSSL}{crystal denoising self-supervised learning}
\newacronym{CHGNET}{CHGNet}{crystal Hamiltonian graph neural network}


\newacronym{DGL}{DGL}{Deep Graph Library}
\newacronym{DFT}{DFT}{density functional theory}



\newacronym[shortplural={GNs}]{GN}{GN}{graph network}



\newacronym{JHU}{JHU}{%
  Johns Hopkins University%
}



\newacronym{MAE}{MAE}{mean absolute error}
\newacronym{MEGNET}{MEGNet}{MatErials Graph Network}
\newacronym{M3GNET}{M3GNet}{universal material graph with three-body interactions neural network}
\newacronym{MGL}{MatGL}{Materials Graph Library}

\newacronym{ML}{ML}{machine learning}
\newacronym{MLP}{MLP}{multi-layer perceptron}

\newacronym{MP}{MP}{Materials Project}
\newacronym{MSE}{MSE}{mean-squared error}

\newacronym{NLP}{NLP}{natural language processing}
\newacronym{NN}{NN}{neural network}
\newacronym{NOMAD}{NOMAD}{Novel Materials Discovery}

\newacronym{OQMD}{OQMD}{Open Quantum Materials Database}



\newacronym{SL}{SL}{supervised learning}
\newacronym{SSL}{SSL}{self-supervised learning}
\newacronym{SGD}{SGD}{stochastic gradient descent}



\newacronym{UMAP}{UMAP}{uniform manifold approximation and projection}



%% file: main.bbl
\begin{thebibliography}{33}
\providecommand{\natexlab}[1]{#1}
\providecommand{\url}[1]{\texttt{#1}}
\expandafter\ifx\csname urlstyle\endcsname\relax
  \providecommand{\doi}[1]{doi: #1}\else
  \providecommand{\doi}{doi: \begingroup \urlstyle{rm}\Url}\fi

\bibitem[Balestriero et~al.(2023)Balestriero, Ibrahim, Sobal, Morcos, Shekhar, Goldstein, Bordes, Bardes, Mialon, Tian, Schwarzschild, Wilson, Geiping, Garrido, Fernandez, Bar, Pirsiavash, LeCun, and Goldblum]{Balestriero2023sslcookbook}
Balestriero, R., Ibrahim, M., Sobal, V., Morcos, A., Shekhar, S., Goldstein, T., Bordes, F., Bardes, A., Mialon, G., Tian, Y., Schwarzschild, A., Wilson, A.~G., Geiping, J., Garrido, Q., Fernandez, P., Bar, A., Pirsiavash, H., LeCun, Y., and Goldblum, M.
\newblock A cookbook of self-supervised learning, 2023.

\bibitem[Battaglia et~al.(2018)Battaglia, Hamrick, Bapst, Sanchez-Gonzalez, Zambaldi, et~al.]{Battaglia2018graphnetworks}
Battaglia, P.~W., Hamrick, J.~B., Bapst, V., Sanchez-Gonzalez, A., Zambaldi, V., et~al.
\newblock Relational inductive biases, deep learning, and graph networks, 2018.
\newblock URL \url{https://arxiv.org/abs/1806.01261}.

\bibitem[Chen \& Ong(2022)Chen and Ong]{Chen2022m3gnet}
Chen, C. and Ong, S.~P.
\newblock A universal graph deep learning interatomic potential for the periodic table.
\newblock \emph{Nature Computational Science}, 2\penalty0 (11):\penalty0 718--728, Nov 2022.
\newblock ISSN 2662-8457.
\newblock \doi{10.1038/s43588-022-00349-3}.
\newblock URL \url{https://doi.org/10.1038/s43588-022-00349-3}.

\bibitem[Chen et~al.(2019)Chen, Ye, Zuo, Zheng, and Ong]{chen2019megnet}
Chen, C., Ye, W., Zuo, Y., Zheng, C., and Ong, S.~P.
\newblock Graph networks as a universal machine learning framework for molecules and crystals.
\newblock \emph{Chemistry of Materials}, 2019.
\newblock \doi{10.1021/acs.chemmater.9b01294}.
\newblock URL \url{https://doi.org/10.1021/acs.chemmater.9b01294}.

\bibitem[Choudhary \& DeCost(2021)Choudhary and DeCost]{Choudhary2021alignn}
Choudhary, K. and DeCost, B.
\newblock Atomistic line graph neural network for improved materials property predictions.
\newblock \emph{npj Computational Materials}, 7\penalty0 (1):\penalty0 185, Nov 2021.
\newblock ISSN 2057-3960.
\newblock \doi{10.1038/s41524-021-00650-1}.
\newblock URL \url{https://doi.org/10.1038/s41524-021-00650-1}.

\bibitem[Deng et~al.(2023)Deng, Zhong, Jun, Riebesell, Han, Bartel, and Ceder]{Deng2023Chgnet}
Deng, B., Zhong, P., Jun, K., Riebesell, J., Han, K., Bartel, C.~J., and Ceder, G.
\newblock Chgnet as a pretrained universal neural network potential for charge-informed atomistic modelling.
\newblock \emph{Nature Machine Intelligence}, pp.\  1–11, 2023.
\newblock \doi{10.1038/s42256-023-00716-3}.

\bibitem[Dunn et~al.(2020)Dunn, Wang, Ganose, Dopp, and Jain]{Dunn2020matbench}
Dunn, A., Wang, Q., Ganose, A., Dopp, D., and Jain, A.
\newblock Benchmarking materials property prediction methods: the matbench test set and automatminer reference algorithm.
\newblock \emph{npj Computational Materials}, 6\penalty0 (1):\penalty0 138, Sep 2020.
\newblock ISSN 2057-3960.
\newblock \doi{10.1038/s41524-020-00406-3}.
\newblock URL \url{https://doi.org/10.1038/s41524-020-00406-3}.

\bibitem[Godwin et~al.(2022)Godwin, Schaarschmidt, Gaunt, Sanchez-Gonzalez, Rubanova, Veli{\v{c}}kovi{\'c}, Kirkpatrick, and Battaglia]{Godwin2022simple}
Godwin, J., Schaarschmidt, M., Gaunt, A.~L., Sanchez-Gonzalez, A., Rubanova, Y., Veli{\v{c}}kovi{\'c}, P., Kirkpatrick, J., and Battaglia, P.
\newblock Simple {GNN} regularisation for 3d molecular property prediction and beyond.
\newblock In \emph{International Conference on Learning Representations}, 2022.
\newblock URL \url{https://openreview.net/forum?id=1wVvweK3oIb}.

\bibitem[Goodall \& Lee(2020)Goodall and Lee]{Goodall2020roost}
Goodall, R. E.~A. and Lee, A.~A.
\newblock Predicting materials properties without crystal structure: deep representation learning from stoichiometry.
\newblock \emph{Nature Communications}, 11\penalty0 (1):\penalty0 6280, Dec 2020.
\newblock ISSN 2041-1723.
\newblock \doi{10.1038/s41467-020-19964-7}.
\newblock URL \url{https://doi.org/10.1038/s41467-020-19964-7}.

\bibitem[Hu et~al.(2020)Hu, Liu, Gomes, Zitnik, Liang, Pande, and Leskovec]{Hu2020PretrainingStrategies}
Hu, W., Liu, B., Gomes, J., Zitnik, M., Liang, P., Pande, V., and Leskovec, J.
\newblock Strategies for pre-training graph neural networks.
\newblock In \emph{International Conference on Learning Representations}, 2020.
\newblock URL \url{https://openreview.net/forum?id=HJlWWJSFDH}.

\bibitem[Huang et~al.(2024)Huang, Magar, and Barati~Farimani]{Huang2024compositionssl}
Huang, H., Magar, R., and Barati~Farimani, A.
\newblock Pretraining strategies for structure agnostic material property prediction.
\newblock \emph{Journal of Chemical Information and Modeling}, 64\penalty0 (3):\penalty0 627--637, 2024.
\newblock \doi{10.1021/acs.jcim.3c00919}.
\newblock URL \url{https://doi.org/10.1021/acs.jcim.3c00919}.
\newblock PMID: 38301621.

\bibitem[Kingma \& Ba(2014)Kingma and Ba]{Kingma2014adam}
Kingma, D.~P. and Ba, J.
\newblock Adam: A method for stochastic optimization, 2014.
\newblock URL \url{https://arxiv.org/abs/1412.6980}.

\bibitem[Kirklin et~al.(2015)Kirklin, Saal, Meredig, Thompson, Doak, Aykol, R{\"u}hl, and Wolverton]{Kirklin2015OQMD}
Kirklin, S., Saal, J.~E., Meredig, B., Thompson, A., Doak, J.~W., Aykol, M., R{\"u}hl, S., and Wolverton, C.
\newblock The {O}pen {Q}uantum {M}aterials {D}atabase ({OQMD}): assessing the accuracy of {DFT} formation energies.
\newblock \emph{npj Computational Materials}, 1\penalty0 (1):\penalty0 15010, Dec 2015.
\newblock ISSN 2057-3960.
\newblock \doi{10.1038/npjcompumats.2015.10}.
\newblock URL \url{https://doi.org/10.1038/npjcompumats.2015.10}.

\bibitem[Ko et~al.(2021)Ko, Nassar, Miret, Liu, Qi, and Ong]{Ko2021matgl}
Ko, T.~W., Nassar, M., Miret, S., Liu, E., Qi, J., and Ong, S.~P.
\newblock Materials graph library, 2021.

\bibitem[Magar et~al.(2022)Magar, Wang, and Barati~Farimani]{Magar2022crystaltwin}
Magar, R., Wang, Y., and Barati~Farimani, A.
\newblock Crystal twins: self-supervised learning for crystalline material property prediction.
\newblock \emph{npj Computational Materials}, 8\penalty0 (1):\penalty0 231, Nov 2022.
\newblock ISSN 2057-3960.
\newblock \doi{10.1038/s41524-022-00921-5}.
\newblock URL \url{https://doi.org/10.1038/s41524-022-00921-5}.

\bibitem[McInnes et~al.(2020)McInnes, Healy, and Melville]{Mcinnes2020umap}
McInnes, L., Healy, J., and Melville, J.
\newblock Umap: Uniform manifold approximation and projection for dimension reduction, 2020.

\bibitem[New et~al.(2022)New, Pekala, Le, Domenico, Piatko, and Stiles]{New2022curvatureinformed}
New, A., Pekala, M.~J., Le, N.~Q., Domenico, J., Piatko, C.~D., and Stiles, C.~D.
\newblock Curvature-informed multi-task learning for graph networks.
\newblock In \emph{ICML 2022 2nd AI for Science Workshop}, 2022.
\newblock URL \url{https://openreview.net/forum?id=m5RYtApKFOg}.

\bibitem[New et~al.(2023)New, Pekala, Pogue, Le, Domenico, Piatko, and Stiles]{new2023evaluating}
New, A., Pekala, M., Pogue, E.~A., Le, N.~Q., Domenico, J., Piatko, C.~D., and Stiles, C.~D.
\newblock Evaluating the diversity and utility of materials proposed by generative models.
\newblock In \emph{1st Workshop on the Synergy of Scientific and Machine Learning Modeling @ ICML2023}, 2023.
\newblock URL \url{https://openreview.net/forum?id=2ZYbmYTKoR}.

\bibitem[Ong et~al.(2013)Ong, Richards, Jain, Hautier, Kocher, Cholia, Gunter, Chevrier, Persson, and Ceder]{Ong2013pymatgen}
Ong, S.~P., Richards, W.~D., Jain, A., Hautier, G., Kocher, M., Cholia, S., Gunter, D., Chevrier, V.~L., Persson, K.~A., and Ceder, G.
\newblock Python materials genomics (pymatgen): A robust, open-source python library for materials analysis.
\newblock \emph{Computational Materials Science}, 68:\penalty0 314--319, 2013.
\newblock ISSN 0927-0256.
\newblock \doi{https://doi.org/10.1016/j.commatsci.2012.10.028}.
\newblock URL \url{https://www.sciencedirect.com/science/article/pii/S0927025612006295}.

\bibitem[Pogue et~al.(2023)Pogue, New, McElroy, Le, Pekala, McCue, Gienger, Domenico, Hedrick, McQueen, Wilfong, Piatko, Ratto, Lennon, Chung, Montalbano, Bassen, and Stiles]{Pogue2023closedloop}
Pogue, E.~A., New, A., McElroy, K., Le, N.~Q., Pekala, M.~J., McCue, I., Gienger, E., Domenico, J., Hedrick, E., McQueen, T.~M., Wilfong, B., Piatko, C.~D., Ratto, C.~R., Lennon, A., Chung, C., Montalbano, T., Bassen, G., and Stiles, C.~D.
\newblock Closed-loop superconducting materials discovery.
\newblock \emph{npj Computational Materials}, 9\penalty0 (1):\penalty0 181, Oct 2023.
\newblock ISSN 2057-3960.
\newblock \doi{10.1038/s41524-023-01131-3}.
\newblock URL \url{https://doi.org/10.1038/s41524-023-01131-3}.

\bibitem[Ruff et~al.(2024)Ruff, Reiser, Stühmer, and Friederich]{Ruff2024cogn}
Ruff, R., Reiser, P., Stühmer, J., and Friederich, P.
\newblock Connectivity optimized nested line graph networks for crystal structures.
\newblock \emph{Digital Discovery}, 3:\penalty0 594--601, 2024.
\newblock \doi{10.1039/D4DD00018H}.
\newblock URL \url{http://dx.doi.org/10.1039/D4DD00018H}.

\bibitem[Scheidgen et~al.(2023)Scheidgen, Himanen, Ladines, Sikter, Nakhaee, Ádám Fekete, Chang, Golparvar, Márquez, Brockhauser, Brückner, Ghiringhelli, Dietrich, Lehmberg, Denell, Albino, Näsström, Shabih, Dobener, Kühbach, Mozumder, Rudzinski, Daelman, Pizarro, Kuban, Salazar, Ondračka, Bungartz, and Draxl]{Scheidgen2023nomad}
Scheidgen, M., Himanen, L., Ladines, A.~N., Sikter, D., Nakhaee, M., Ádám Fekete, Chang, T., Golparvar, A., Márquez, J.~A., Brockhauser, S., Brückner, S., Ghiringhelli, L.~M., Dietrich, F., Lehmberg, D., Denell, T., Albino, A., Näsström, H., Shabih, S., Dobener, F., Kühbach, M., Mozumder, R., Rudzinski, J.~F., Daelman, N., Pizarro, J.~M., Kuban, M., Salazar, C., Ondračka, P., Bungartz, H.-J., and Draxl, C.
\newblock Nomad: A distributed web-based platform for managing materials science research data.
\newblock \emph{Journal of Open Source Software}, 8\penalty0 (90):\penalty0 5388, 2023.
\newblock \doi{10.21105/joss.05388}.
\newblock URL \url{https://doi.org/10.21105/joss.05388}.

\bibitem[Vinyals et~al.(2016)Vinyals, Bengio, and Kudlur]{Vinyals2016set2set}
Vinyals, O., Bengio, S., and Kudlur, M.
\newblock Order matters: Sequence to sequence for sets, 2016.

\bibitem[Wang et~al.(2021)Wang, Kauwe, Murdock, and Sparks]{Wang2021crabnet}
Wang, A. Y.-T., Kauwe, S.~K., Murdock, R.~J., and Sparks, T.~D.
\newblock Compositionally restricted attention-based network for materials property predictions.
\newblock \emph{npj Computational Materials}, 7\penalty0 (1):\penalty0 77, May 2021.
\newblock ISSN 2057-3960.
\newblock \doi{10.1038/s41524-021-00545-1}.
\newblock URL \url{https://doi.org/10.1038/s41524-021-00545-1}.

\bibitem[Wang et~al.(2020)Wang, Zheng, Ye, Gan, Li, Song, Zhou, Ma, Yu, Gai, Xiao, He, Karypis, Li, and Zhang]{Wang2020dgl}
Wang, M., Zheng, D., Ye, Z., Gan, Q., Li, M., Song, X., Zhou, J., Ma, C., Yu, L., Gai, Y., Xiao, T., He, T., Karypis, G., Li, J., and Zhang, Z.
\newblock Deep graph library: A graph-centric, highly-performant package for graph neural networks, 2020.

\bibitem[Ward et~al.(2016)Ward, Agrawal, Choudhary, and Wolverton]{Ward2016magpie}
Ward, L., Agrawal, A., Choudhary, A., and Wolverton, C.
\newblock A general-purpose machine learning framework for predicting properties of inorganic materials.
\newblock \emph{npj Computational Materials}, 2\penalty0 (1):\penalty0 16028, Aug 2016.
\newblock ISSN 2057-3960.
\newblock \doi{10.1038/npjcompumats.2016.28}.
\newblock URL \url{https://doi.org/10.1038/npjcompumats.2016.28}.

\bibitem[Ward et~al.(2018)Ward, Dunn, Faghaninia, Zimmermann, Bajaj, Wang, Montoya, Chen, Bystrom, Dylla, Chard, Asta, Persson, Snyder, Foster, and Jain]{Ward2018matminer}
Ward, L., Dunn, A., Faghaninia, A., Zimmermann, N.~E., Bajaj, S., Wang, Q., Montoya, J., Chen, J., Bystrom, K., Dylla, M., Chard, K., Asta, M., Persson, K.~A., Snyder, G.~J., Foster, I., and Jain, A.
\newblock Matminer: An open source toolkit for materials data mining.
\newblock \emph{Computational Materials Science}, 152:\penalty0 60--69, 2018.
\newblock ISSN 0927-0256.
\newblock \doi{https://doi.org/10.1016/j.commatsci.2018.05.018}.
\newblock URL \url{https://www.sciencedirect.com/science/article/pii/S0927025618303252}.

\bibitem[Xie \& Grossman(2018)Xie and Grossman]{Xie2018CGCNN}
Xie, T. and Grossman, J.~C.
\newblock Crystal graph convolutional neural networks for an accurate and interpretable prediction of material properties.
\newblock \emph{Phys. Rev. Lett.}, 120:\penalty0 145301, Apr 2018.
\newblock \doi{10.1103/PhysRevLett.120.145301}.

\bibitem[Xie et~al.(2022)Xie, Fu, Ganea, Barzilay, and Jaakkola]{Xie2022cdvae}
Xie, T., Fu, X., Ganea, O.-E., Barzilay, R., and Jaakkola, T.~S.
\newblock Crystal diffusion variational autoencoder for periodic material generation.
\newblock In \emph{International Conference on Learning Representations}, 2022.
\newblock URL \url{https://openreview.net/forum?id=03RLpj-tc_}.

\bibitem[Xie et~al.(2023)Xie, Xu, Zhang, Wang, and Ji]{Xie2023graphsslreview}
Xie, Y., Xu, Z., Zhang, J., Wang, Z., and Ji, S.
\newblock Self-supervised learning of graph neural networks: A unified review.
\newblock \emph{IEEE Transactions on Pattern Analysis and Machine Intelligence}, 45\penalty0 (2):\penalty0 2412--2429, 2023.
\newblock \doi{10.1109/TPAMI.2022.3170559}.

\bibitem[Zaidi et~al.(2023)Zaidi, Schaarschmidt, Martens, Kim, Teh, Sanchez-Gonzalez, Battaglia, Pascanu, and Godwin]{Zaidi2023denoising}
Zaidi, S., Schaarschmidt, M., Martens, J., Kim, H., Teh, Y.~W., Sanchez-Gonzalez, A., Battaglia, P., Pascanu, R., and Godwin, J.
\newblock Pre-training via denoising for molecular property prediction.
\newblock In \emph{The Eleventh International Conference on Learning Representations}, 2023.
\newblock URL \url{https://openreview.net/forum?id=tYIMtogyee}.

\bibitem[Zeni et~al.(2024)Zeni, Pinsler, Zügner, Fowler, Horton, Fu, Shysheya, Crabbé, Sun, Smith, Nguyen, Schulz, Lewis, Huang, Lu, Zhou, Yang, Hao, Li, Tomioka, and Xie]{Zeni2024mattergen}
Zeni, C., Pinsler, R., Zügner, D., Fowler, A., Horton, M., Fu, X., Shysheya, S., Crabbé, J., Sun, L., Smith, J., Nguyen, B., Schulz, H., Lewis, S., Huang, C.-W., Lu, Z., Zhou, Y., Yang, H., Hao, H., Li, J., Tomioka, R., and Xie, T.
\newblock Mattergen: a generative model for inorganic materials design, 2024.

\bibitem[Zhao et~al.(2023)Zhao, Siriwardane, Wu, Fu, Al-Fahdi, Hu, and Hu]{Zhao2023pgcgm}
Zhao, Y., Siriwardane, E. M.~D., Wu, Z., Fu, N., Al-Fahdi, M., Hu, M., and Hu, J.
\newblock Physics guided deep learning for generative design of crystal materials with symmetry constraints.
\newblock \emph{npj Computational Materials}, 9\penalty0 (1):\penalty0 38, Mar 2023.
\newblock ISSN 2057-3960.
\newblock \doi{10.1038/s41524-023-00987-9}.
\newblock URL \url{https://doi.org/10.1038/s41524-023-00987-9}.

\end{thebibliography}
